\documentclass[conference]{IEEEtran}
\IEEEoverridecommandlockouts
\usepackage{amsmath,amssymb,amsfonts}
\usepackage{algorithmic}
\usepackage{graphicx}
\graphicspath{{./figs/}}
\usepackage{subcaption}
\usepackage{textcomp}
\usepackage{xcolor}
\def\BibTeX{{\rm B\kern-.05em{\sc i\kern-.025em b}\kern-.08em
    T\kern-.1667em\lower.7ex\hbox{E}\kern-.125emX}}
\usepackage[%
    style=ieee,
    citestyle=numeric-comp,
    sortcites=true,
    natbib=true,
    backend=biber,
    maxnames=6,
]{biblatex}

\AtBeginBibliography{\small}
\usepackage{etoolbox}
\apptocmd{\sloppy}{\hbadness 10000\relax}{}{}

\usepackage{multirow}
\usepackage{booktabs}
\usepackage{siunitx}

\begin{document}

\title{Robust Reinforcement Learning-based Autonomous Driving Agent for Simulation and Real World\\
%{\footnotesize %\textsuperscript{*}Note: Sub-titles are not captured in Xplore andshould not be used}
%\thanks{TBD: ACKNOWLEDGEMENT! Projektek, Conti PIA} %TODO
}

\author{\IEEEauthorblockN{P{\'e}ter Alm{\'a}si, R{\'o}bert Moni, B{\'a}lint Gyires-T{\'o}th}
\IEEEauthorblockA{\textit{Department of Telecommunications and Media Informatics} \\
\textit{Budapest University of Technology and Economics}, Budapest, HUNGARY \\
peter.almasi@cs.bme.hu, \{robertmoni,toth.b\}@tmit.bme.hu}
}
\maketitle

\begin{abstract}
Deep Reinforcement Learning (DRL) has been successfully used to solve different challenges, e.g. complex board and computer games, recently. However, solving real-world robotics tasks with DRL seems to be a more difficult challenge. The desired approach would be to train the agent in a simulator and transfer it to the real world. Still, models trained in a simulator tend to perform poorly in real-world environments due to the differences. In this paper, we present a DRL-based algorithm that is capable of performing autonomous robot control using Deep Q-Networks (DQN). In our approach, the agent is trained in a simulated environment and it is able to navigate both in a simulated and real-world environment. The method is evaluated in the Duckietown environment, where the agent has to follow the lane based on a monocular camera input. The trained agent is able to run on limited hardware resources and its performance is comparable to state-of-the-art approaches. 
\end{abstract}


\section{Introduction}

    Artificial Intelligence (AI) has become a very focused research area in the past years. Among its subfields, deep learning has been one of the most important ones due to the state-of-the-art results reached in several application scenarios, e.g. image recognition, speech recognition and synthesis, natural language processing and reinforcement learning.
    
	Deep learning has an important role in autonomous driving. The development of deep neural networks and the supporting hardware and software solutions made it possible to build robust computer vision models. Deep convolutional neural networks \cite{lecun1998gradient} are outstanding in object detection and semantic segmentation \cite{NIPS2015_5638} \cite{chen2018encoder}. I.e. it is possible to find and localize certain objects (e.g. cars, pedestrians, cyclists or traffic signs, etc.) on the images, which is critical for autonomous vehicles. 
	%Finding and localizing objects around the autonomous vehicle can help making driving decisions in order to avoid collisions.
	
	A further method described in \cite{bojarski2016end} trains a convolutional neural network (CNN) to map raw images of the camera directly to steering commands. They use an end-to-end approach, eliminating explicit image preprocessing and object detection steps. This can lead to better performance since the system is trained to maximize overall performance instead of finding manually selected features, e.g. lane detection. \cite{chen2015deepdriving} uses CNNs to predict simple indicators, that describe the actual road situation (e.g. distance from lane markings and other vehicles, the angle relative to the road). With these indicators, they utilize a controller that can make driving decisions at a high-level. The indicators are more compact than, for example, creating segmentation for the image. 
	
	Reinforcement learning (RL) is an area of machine learning where an agent is optimized to take actions in an environment to reach a specified goal, which is represented as a scalar value (called reward). Utilizing deep neural networks in RL enables the agent to be able to learn complex contexts and short and long-term strategies.
	Recent advances in RL made it possible to achieve superhuman level in complex board and computer games, such as Go \cite{silver2017mastering}. \cite{mnih2015human} shows an example of an agent, that was trained on a human-level performance to play many kinds of Atari games (e.g. Breakout) by using high-dimensional images as input only. RL agents are also capable of surpassing human players in more complex computer games: the AlphaStar \cite{vinyals2019grandmaster} successfully overcome a professional player in the StarCraft II game, and the OpenAI Five \cite{berner2019dota} can win against human players in the Dota 2 game.
	
	The application of DRL algorithms in autonomous vehicles is currently in an initial phase. Training agents for real-world problems in a simulator is a promising approach, as it is much safer to simulate incidences that must be avoided in the real world (e.g. collisions, running over pedestrians). Also, with sufficient GPU resources, the agents can be trained in a much faster pace than real-time. Collecting sufficient training data is also much more convenient within a simulator. The simulator can also provide additional metrics (e.g. accurate distance between objects and their location) which may be difficult to be measured in the real world but help to evaluate the performance of the agent. However, simulators often have significant differences compared to the real world, and these differences (e.g. details, colors, lighting conditions, or dynamics) can cause the trained models to suffer significant performance degradation in the real world. Training autonomous vehicle agents in simulators with reinforcement learning and transferring the agents to the real world are both active research areas that are in the early stages. 
	%Thus, it is important to examine the possibilities of solving these challenges in a simplified environment with basic methods, first.
	
    This paper presents a reinforcement learning pipeline for training an agent in an autonomous driving simulator and running the trained agent in the real world. The rest of this paper is organized as follows. Section \ref{sec:background} gives an introduction to reinforcement learning and domain randomization, Section \ref{sec:training} describes the details of the proposed method, Section \ref{sec:environment} presents the Duckietown environment in which the proposed method was tested. Evaluation and results are introduced in Section \ref{sec:results}, and conclusions are drawn in Section \ref{sec:conclusion}.

\section{Background}
\label{sec:background}

\subsection{Reinforcement learning}

In this paper, we focus on basic RL settings where an agent interacts with an environment by following a policy in order to maximize a reward. There are two main approaches of RL settings: model-based and model-free methods.

Model-based methods construct an internal model of the environment by experiencing transitions and learning the dynamics of the environment. Using this internal model actions are taken either by searching or planning in this world model. 

Model-free methods also rely on prior experiences and aim to learn the state-action values, or policy, or both of them. These are the: value learning, policy optimization and actor-critic methods respectively. Value learning methods focus on estimating the value of being in a given state. Policy optimization methods focus on finding the optimal policy that maximizes the expected return, also called reward. The actor-critic method is a hybrid method where a policy optimization method ("actor") is learning from the feedback of the value function ("critic").

Each of these methods are constructed to fit the Markov Decision Process (MDP). We selected the model-free value learning (Q learning) method, which tailored to our setup works as follows: in each timestep $t$, the environment provides the agent its state $s_t$ in the form of a high-dimensional RGB image. The agent chooses an action $a_t$ from a given set of possible actions according to a policy $\pi(a|s)$. The environment computes a reward $r_t$ based on the chosen action at the given state which describes how 'good' the selected action was. The agent tries to maximize the total reward $R_t=\sum_{u=t}^{t+T}{r_u\gamma^{u-t}}$, where $\gamma$ is the discount factor ($\gamma\in[0,1]$), and $T$ is the length of the episode. The goal during learning is to find an optimal policy $\pi^*$ which maximizes the expected reward the agent receives in each episode. 

Since the observed states of the environment are represented by RGB images and the action space is discrete, we chose the model-free and off-policy Deep Q-Networks \cite{mnih2013playing} \cite{mnih2015human} method with experience replay. Despite a basic Deep Reinforcement Learning method is utilized, it is capable of learning an optimal policy in simulated traffic environment which performs nearly as good in the real environment by using proper domain randomization methods. 
In this work, we propose a CNN + DQN system that can be trained with low expenses in a simulated traffic environment and can be transferred to a robot to perform autonomous driving in a miniaturized urban environment. 

%\subsection{Deep Q-Networks}
%Deep Q-Networks (DQN) \cite{mnih2013playing} \cite{mnih2015human} is a reinforcement learning algorithm that was originally developed to successfully play Atari games by processing high-dimensional visional input. We used this algorithm in our experiments, as in our task the agent also receives images as the current state, and has to take actions based on it by understanding and finding the relevant parts of the image. DQN uses a convolutional neural network to approximate the optimal action-value function. An experience replay buffer is used to store the previous experiences of the agent; updating the weights of the neural network based on samples from this buffer (instead of only the current one) helps to make the training more stable. The algorithm is model-free and off-policy.

\subsection{Reinforcement Learning in Autonomous Driving}
The new era of Advanced Driving Assistance Systems (ADASs) tend to apply deep learning methods for scene perception, object localization, path planning, behavior arbitration, and motion control \cite{autsurvey2019}. These system gather information from the environment using cameras and localization systems such as RADAR and Li-DAR. Scene perception, object localization, and behavior arbitration are generally supervised learning problems where the training requires large amounts of labeled data. Path planning and motion control can also be tackled using supervised learning methods. One of the first deep learning based autonomous driving solution was ALVINN \cite{Pomerleau1988ALVINNAA} published in 1989 consisting of a simple 3-layered neural network trained with simulated road images. \cite{bechtel2018deepicar}, \cite{bojarski2017pilotnet} and \cite{agile2018imit} applies imitation learning technique to train the agent with a CNN architecture using monocular camera images and recorded human-driver speed and steering angles. This method tends to learn promptly the expert's (human driver) behavior, but performs poorly in new environments. Also the producing, labeling and storing of the training data is both time inefficient and expensive. \cite{sallab2017torcs} proposes a CNN+RNN+DQN architecture which performs well in a car racing simulated environment, but the environment is observed from a top-view which is not adequate for real-world scenarios. 
   
    \begin{figure*}
		\centering
		\includegraphics[width=0.8\textwidth]{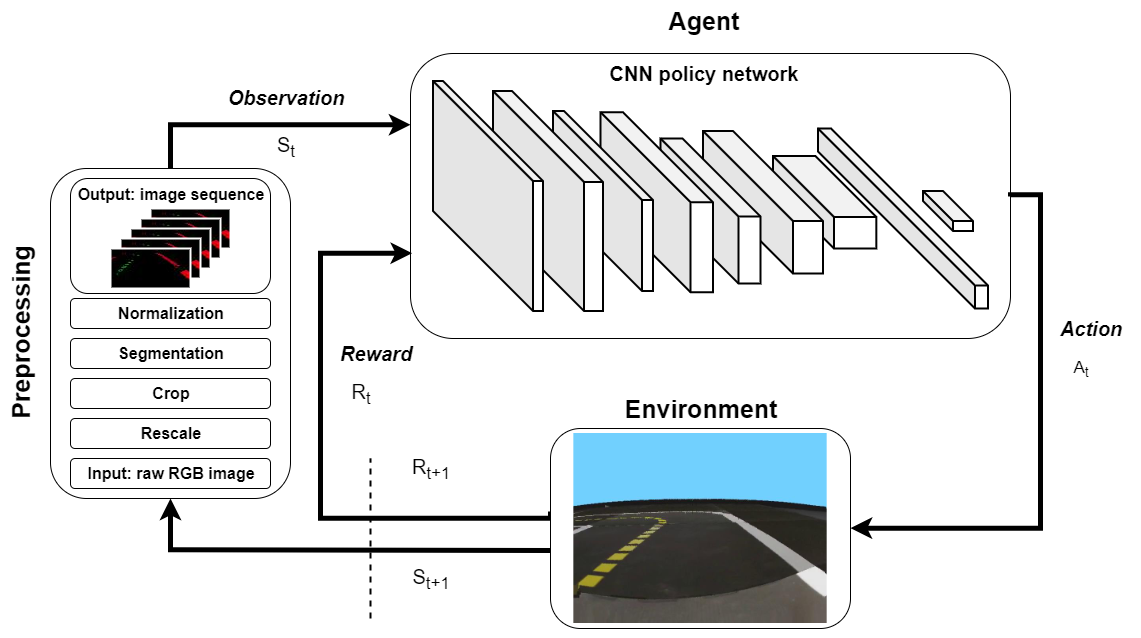}
		\caption{Overview of our method.}
		\label{fig:pipeline}
	\end{figure*}

\subsection{Domain randomization}

Training or transferring robot control agents to the real world is challenging. Domain randomization improves the agent in the training phase in the simulator to have similar performance in the real world. \cite{tobin2017domain} uses this method for object localization on real-world images by training a neural network in a simulator with generated images. The images are highly manipulated (e.g. objects' position, textures, camera position, lighting conditions, etc.) and these manipulated data are used to train the network in the simulator. \cite{tremblay2018training} also utilizes object localization with images generated in a simulator and modified using domain randomization. They show that their method is comparable to the case of training on manually annotated real-world images. \cite{peng2018sim} uses domain randomization to generate different dynamics in the simulator, and train a neural network which makes a robotic arm able to move objects to an assigned location. They show that training in the simulator with randomized dynamics makes the robotic arm in the real world able to work without any further training on the physical system. \cite{chebotar2019closing} solves the problem of transferring models to real-world robots by adapting simulation randomization using real-world data to learn simulator parameter distributions that are suitable for a successful policy transfer. They change the distribution of the simulations in iterative steps to improve policy transfer by matching the policy behavior in simulation and the real world. \cite{akkaya2019solving} introduces Automatic Domain Randomization, which utilizes incrementally challenging environments in the simulation. They used this technique to train a robotic arm to solve Rubik's cube. Starting with a single, non-randomized environment, the amount of domain randomization is regularly increased as the model learns to perform well in the previous environments. This way the neural network learns to generalize in randomized environments and becomes able to solve the task in difficult conditions, thus making it possible to successfully transfer the model trained in a simulator to the physical robot.

Domain randomization includes diversified technologies to help to transfer the agents trained in the simulator to the real world. Currently, there is no common solution for autonomous driving. Hence, it was our motivation to develop a pipeline to train RL agents in simulators and run them in the real world without severe degradation in performance.

\section{Proposed Method}
\label{sec:training}
    
In this section, we present the details of the proposed method. Fig. \ref{fig:pipeline} shows an overview of the proposed pipeline. First, the camera images go through several preprocessing steps. Next, a sequence is formed from the last $k$ camera images, which will be the input of the CNN policy network (the agent). The agent is trained in the simulator with the DQN algorithm based on the reward. The output of the network is mapped to wheel speed commands. 
%We suppose that the interval of the commands in the simulator and in reality is the same. 

\subsection{Image preprocessing}

Before the images of the simulator's monocular virtual camera are fed to the neural networks, we preprocess them. In the case of the real-world monocular camera, the same steps are performed.

    \begin{itemize}
		\item \textbf{Resizing}: The images are downscaled from their original size (e.g. $640 \times 480$) to a smaller resolution (e.g. $80 \times 60$). This step makes training the neural networks and inference faster, while the smaller resolution image still provides enough information for navigating the robot.
		
		\item \textbf{Cropping}: The part of the image that doesn't contain useful information is cropped. This is typically done above the horizon.
		
		\item \textbf{Color segmentation}: To make it easier for the neural network to recognize the important parts of the image, the key colors are segmented based on their values. 
		The original image's channels are substituted with the segmented parts. 
		E.g. for lane following, the yellow and the white parts of the image (that define the two lanes) are separated in the image's red and green channels, respectively. 
		
		\item \textbf{Normalization}: The pixel values are normalized to the $[0, 1]$ range, which helps to train the CNN faster.
		
		\item \textbf{Image sequence}: The last $k$ camera images are concatenated into a 3D tensor with dimensions ($height$, $width$, $sequence\_length \times channels$). This way a time series is formed from the previous states of the environment, which provides richer information and results in better policy networks, compared to a single instance. 
		
	\end{itemize}
	
\begin{figure}[tb]
		\centering
		\begin{subfigure}{0.2\textwidth}
			\centering
			\includegraphics[height=2.5cm]{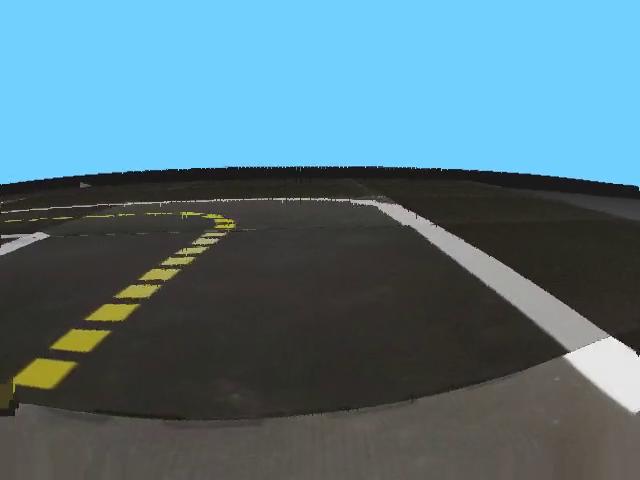}
			%		\caption{The Duckietown platform.}
			\label{fig:fig4a-sim-orig}
		\end{subfigure}
		\begin{subfigure}{0.2\textwidth}
			\centering
			\includegraphics[height=1.7cm]{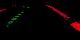}
			%		\caption{Duckiebot}
			\label{fig:fig4b-sim-prep}
		\end{subfigure}
		\caption{Original image in the simulator (left) and after preprocessing it (right).}
		\label{fig:fig4-image-preprocess}
	\end{figure}

	An example of an original and a preprocessed single image is shown in Fig. \ref{fig:fig4-image-preprocess}.

\subsection{Policy network}

\begin{figure}[tb]
		\centering
		\includegraphics[height=5.6cm]{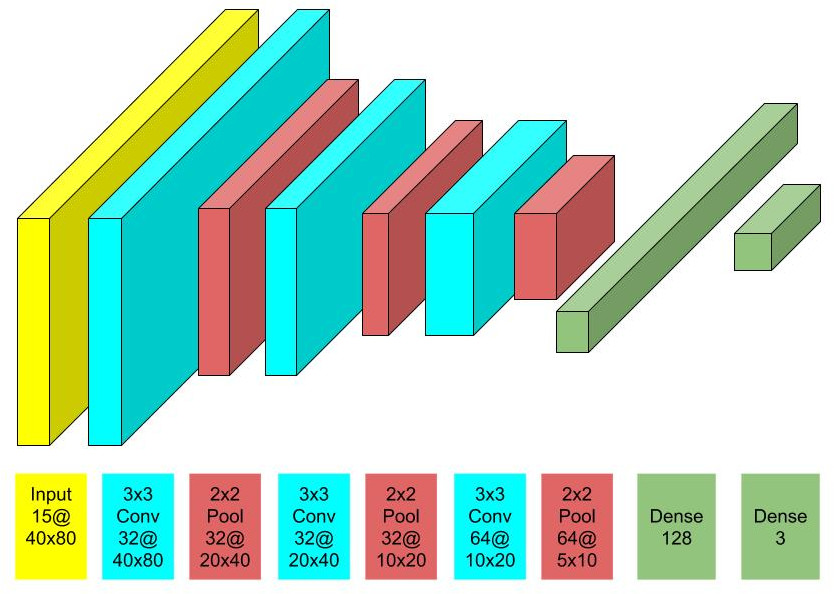}
		\caption{The policy network we used for modeling the input image sequences.}
		\label{fig:nn}
	\end{figure}

We trained a convolutional neural network with the preprocessed images. The network was designed such that the inference can be be performed real-time on a computer with limited resources. The input of the network is a tensor with the shape of the image sequence ($height$, $width$, $sequence\_length \times channels$), e.g. ($40, 80, 15$), which is the result of stacking five RGB images. For fast inference and for demonstration purposes we utilized a simple neural network. The neural network consists of three convolutional layers, each followed by ReLU (nonlinear activation function) and MaxPool (dimension reduction) operations. The convolutional layers use $32, 32, 64$ filters with size $3\times3$. The MaxPool layers use $2\times2$ filters, so they reduce the size of their input to its $1/4$-th. The convolutional layers are followed by fully connected layers with $128$ and $3$ outputs. The output of the last layer corresponds to the selected action. The architecture of the policy network is shown in Fig. \ref{fig:nn}. In the case of more complex environments, the size of the policy network can be increased, indeed.
	
The output of the neural network (one of the three actions) is mapped to wheel speed commands. The actions correspond to turning left, turning right or going straight, respectively.

\section{Environment}
\label{sec:environment}

\begin{figure}[tb]
		\centering
		\begin{subfigure}{0.4\textwidth}
			\centering
			\includegraphics[height=4cm]{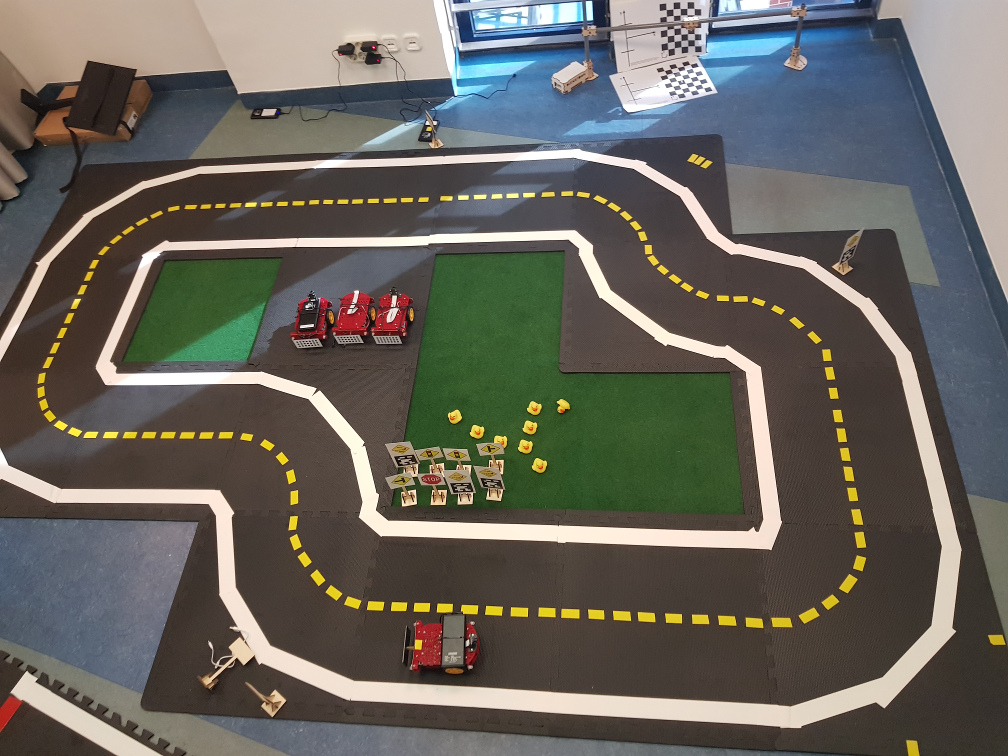}
			\caption{An example setup for the Duckietown Platform.}
			\label{fig:fig-platform}
		\end{subfigure}
		
		\begin{subfigure}{0.4\textwidth}
			\centering
			\includegraphics[height=4cm]{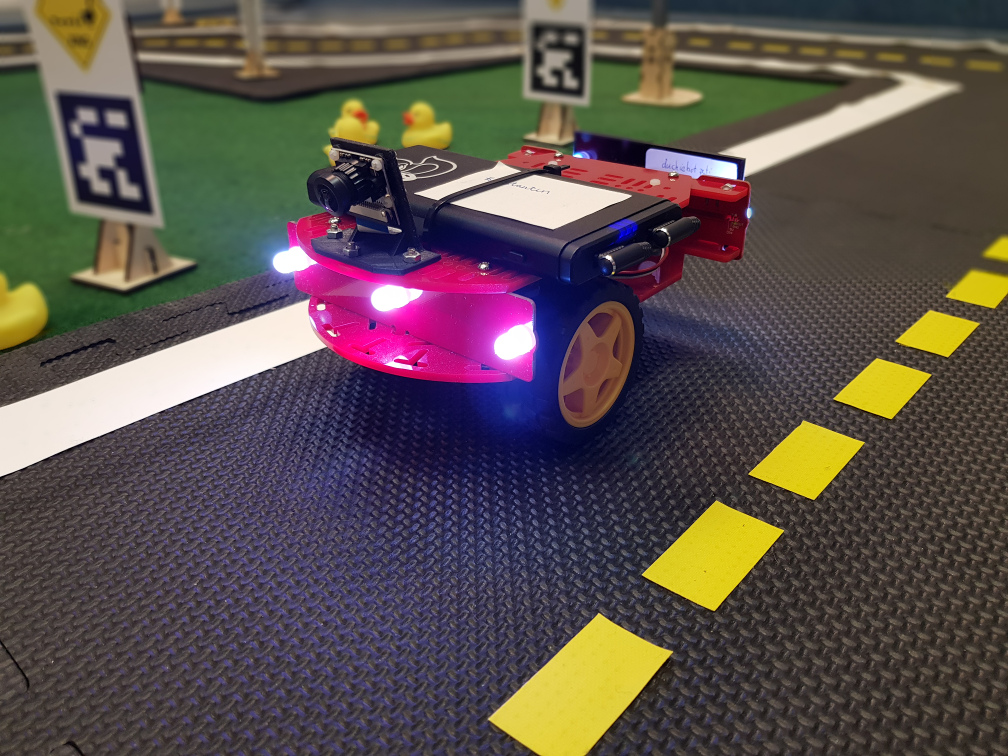}
			\caption{Duckiebot}
			\label{fig:fig-car}
		\end{subfigure}
		\caption{The Duckietown platform and the Duckiebot.}
		\label{fig:fig-duckie}

	\end{figure}

We used the Duckietown\footnote{\url{https://www.duckietown.org/}} environment for evaluation. Duckietown is an educational and research platform where low-cost robots ('Duckiebots') can travel in a small city ('Duckietown') \cite{paull2017duckietown}. The Duckietown and the Duckiebot are shown in Fig. \ref{fig:fig-duckie}. The Duckiebots are small three-wheeled vehicles built almost entirely from off-the-shelf parts. They have only one sensor: a forward-facing wide-angle monocular camera, which they can use to get information about the surrounding objects. The computation is performed by a Raspberry Pi 3, which is responsible for getting the images and controlling the robot. Duckietowns are the cities where the Duckiebots have to operate. These consist of roads, intersections, traffic signs, houses, rubber ducks and other obstacles. The platform is highly flexible: using the standardized road elements, different kinds of cities can be built.
	
	A general goal of the Duckietown project is to provide an environment similar to a real-world autonomous driving environment for a much lower price, which makes it available for educational and research purposes for a wider range of researchers. While being much cheaper, the environment provides similar challenges to those that are accessible in a more complex autonomous driving platform.
	
    \subsubsection{Duckietown Simulator}
	
	The Duckietown software library contains a Duckietown Simulator \cite{gym_duckietown}. The Simulator provides a similar environment to the real-world Duckietowns: it simulates roads, intersections, obstacles (e.g. vehicles or duckies on the road) and other Duckiebots. Using the simulator, it is possible to manually drive an agent around the map or to test how the trained agent can navigate. The simulator places the robot onto a given map and generates the image that the robot's camera would see in a real-world environment. The robot can be controlled by specifying the speeds of the wheels (two values between -1 and 1 for the two wheels).
	
    \subsubsection{Real-world Duckietown}
    
    After training the models in the Duckietown simulator, we tested the trained agent in a real-world environment, which is shown in Fig. \ref{fig:fig-platform}.

\begin{figure}[tb]
		\centering
		\begin{subfigure}{0.2\textwidth}
			\centering
			\includegraphics[height=2.7cm]{407-sim.jpg}
		\end{subfigure}
		\begin{subfigure}{0.2\textwidth}
			\centering
 			\includegraphics[height=2.7cm]{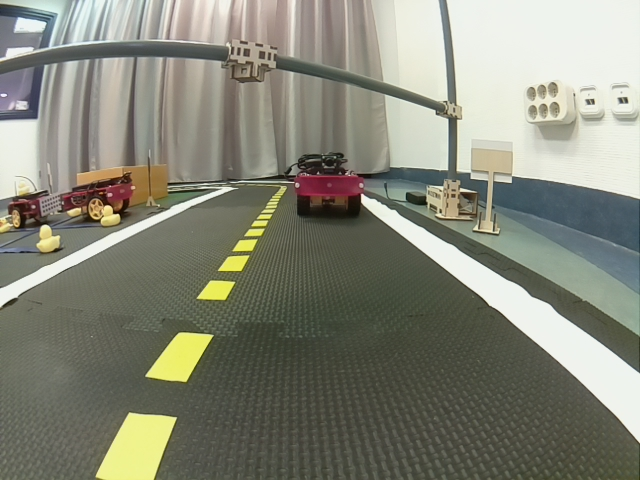}
		\end{subfigure}
		\caption{Images from the Duckietown simulator (left) and the camera of the robot (right). The robot's only sensor is its camera, so it has to be controlled based only this information. Notice that the real-world image has different lighting conditions and camera angle, which makes transferring the trained models to the real robot more difficult.}
		\label{fig:fig-simulator}
	\end{figure}

    Testing in the real-world environment poses new challenges in addition to the simulator. For example, the real-world images seen by the robots are different than those provided by the simulator (a comparison can be seen in Fig. \ref{fig:fig-simulator}.) It is visible that the simulator images, while being similar to the real ones, has different lighting conditions, camera angle, and has a simpler setup regarding the objects surrounding the track, which makes it harder to transform the agent to the real-world robot. Another challenge that arises when evaluating on the real robot is that the robot has to be controlled in real-time: while in the simulator it is possible to simulate a slower algorithm, in the real world, the camera image must be processed in a few milliseconds, which means that the neural network has to be designed carefully such that it can predict driving commands from the camera images fast. Currently, the images are processed on a x86 CPU (instead of the robot's Raspberry Pi computer), which results in smaller inference time. It is also worth noting that there are other factors which make it harder to evaluate the solutions on the real robot compared to the simulator (e.g. network delays).
    
    \subsubsection{Training details}
    
    We used the DQN implementation available in the Stable Baselines collection \cite{stable-baselines}. The algorithm chooses one of three possible actions at each timestep; the mapping between these actions and the robot wheel speeds can be seen in Table \ref{tab:discrete-actions}. We experimented with different hyperparameter settings, including different values and settings for the learning rate, input image size, experience replay buffer size, image segmentation parameters, camera distortion, discount value, policy network parameters, and wheel speeds. The parameters that gave the best results were the following. We used a batch size of $32$, gamma=$0.99$, the learning rate for the Adam optimizer was set to $0.00005$, the size of the replay buffer was $50000$, and the agent collected 10000 steps of experience from random actions before actually starting learning. We ran the training for $500000$ timesteps, which took approximately 40 hours on an NVIDIA DGX Workstation, which contains 4 pieces of V100 GPUs.

	\begin{table}
	
		\caption{Discrete actions predicted by the DQN algorithm are mapped to wheel speeds. (Maximum speed is 1.0.)}
		
		\begin{center}
		\begin{tabular}{lSS}
			\toprule
			Action & {Left wheel speed} & {Right wheel speed} \\
			\midrule        
			{0 (Left)} & 0.04 & 0.4 \\ 
			%\hline 
			{1 (Right)} & 0.4 & 0.04 \\ 
			%\hline
			{2 (Straight)} & 0.3 & 0.3 \\ 
			\bottomrule
		\end{tabular}
		\label{tab:discrete-actions}
		\end{center}
	\end{table}

The simulator provides a reward function, which can be used for reinforcement learning-based methods. This reward reflects how accurately the agent follows its lane. When the agent is going in the right lane, it receives a positive reward. When it starts to drift away from the optimal curve, it receives smaller rewards; when it goes to the oncoming lane, it receives smaller negative rewards, and it gets penalty when it leaves the track (the simulated episode also ends at this point). 

We slightly modified the default reward provided by the simulator. When the robot is in the right lane, the reward is calculated according to the following formula:
$$
reward = 10 \cdot speed \cdot dot\_dir -100 \cdot dist + 400 \cdot col\_pen,
$$
where $speed$ is the speed of the robot in the simulator, $dot\_dir$ is calculated as the dot product of the vectors pointing towards the heading of the robot and the tangent of the curve, $dist$ is the distance from the center of the right lane, and $col\_pen$ is a penalty for collisions. When the robot is not in the right lane, the reward is:
$$
reward = 400 \cdot col\_pen.
$$
When the robot leaves the track, it gets a reward of $-40$.

\begin{figure}[tb]
		\centering
		\includegraphics[height=5.4cm]{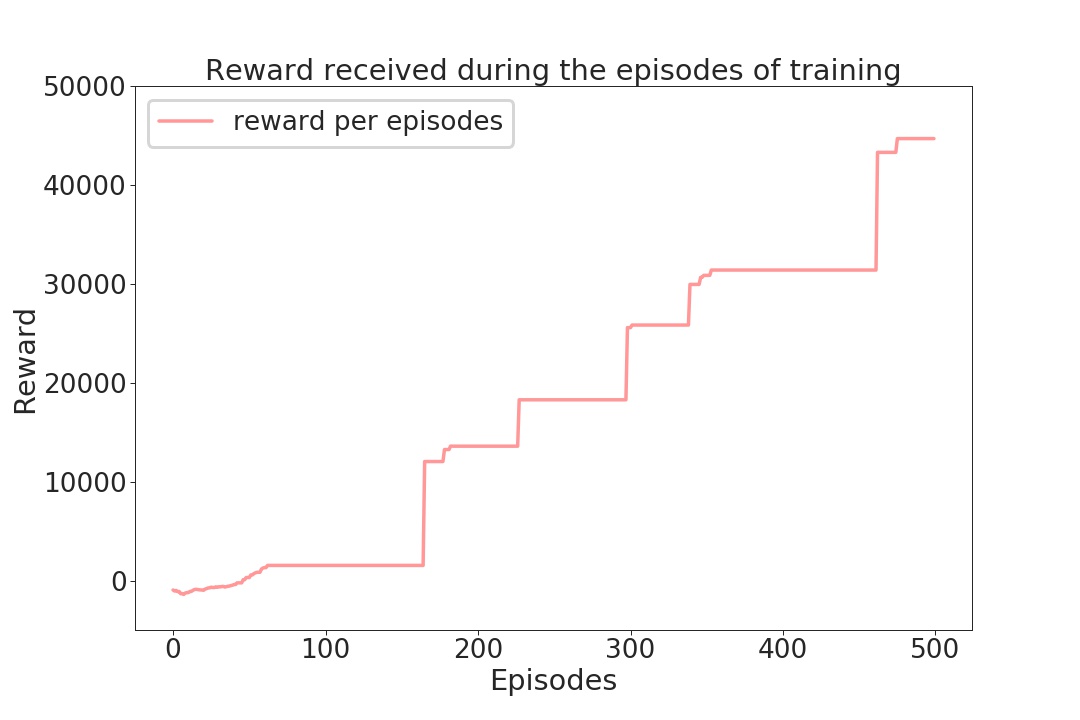}
		\caption{Received rewards during the episodes of the training of the agent.}
		\label{fig:rewards}
	\end{figure}

The rewards achieved throughout the episodes of training phase are shown in Fig. \ref{fig:rewards}.

\section{Results}
\label{sec:results}

Our primary goal was to train an agent in the simulator which can navigate the robot along the track both in the simulator and the real world. We tested our method on several maps, different from the one we used during training, to eliminate the possibility of overfitting to one single map. We trained on a larger, more complicated map, to make it possible for the network to learn diverse turns and road situations. A video of our robot in action can be seen at \url{http://bit.ly/wcci20duckie}.

\subsubsection{Performance in the simulator and in real world}

We used three maps for testing: Map \#1, Map \#2 and Map \#3 can be seen in Figs. \ref{fig:sim-map-a}-\ref{fig:sim-map-f}, \ref{fig:sim-map-g} and \ref{fig:sim-map-h} respectively. The real-world environment is built according to Map \#1 (see Fig. \ref{fig:fig-platform}). We placed the robot on 50 randomly selected positions of the map and counted the number of occasions it was able to drive for a complete lap on the track. We excluded those randomly generated situations from our tests where the robot was dropped to the side of the track facing outwards, where it was impossible to navigate back to the track. In the simulator, we limited the lengths of the episodes for 2500 and 3500 timesteps (approx. 50-70 seconds), which is enough time for the robot to take a whole lap on the maps. In the real-world environment, we ran the evaluation for 45 seconds, which also was enough for completing a full lap. The results of our tests can be seen in Table \ref{tab:results}. We also ran two longer tests in the simulator (50000 timesteps) and found that when the robot was able to take at least one complete lap, it was also able to drive for 50000 timesteps without leaving the track. Therefore, we decided to test our model more thoroughly for only the duration of one complete lap, since after it takes one lap, it has successfully gone through all parts of the track, and we can assume that it could do the same in the following laps too.

\begin{table}[tbp]
\caption{Rates of successful drives on three simulated and one real-world map.}
\begin{center}
\begin{tabular}{lSSS}
\toprule
\multirow{2}{*}{\textbf{Environment}} & \multicolumn{3}{c}{\textbf{Tests}} \\
& \textbf{Total} & \textbf{Successful} & \textbf{Success rate} \\
\midrule
\textbf{Simulator Map \#1} & 50 & 38 & {76\%} \\
%\hline
\textbf{Simulator Map \#2} & 50 & 49 & {98\%} \\
%\hline
\textbf{Simulator Map \#3} & 50 & 41 & {82\%} \\
\midrule
\textbf{Real world} & 50 & 48 & {96\%} \\ 
\bottomrule
\end{tabular}
\label{tab:results}
\end{center}
\end{table}

\subsubsection{Agent navigation patterns}

\begin{figure}[tb]
	\centering
	\begin{subfigure}{0.2\textwidth}
		\centering
		\includegraphics[height=2.3cm]{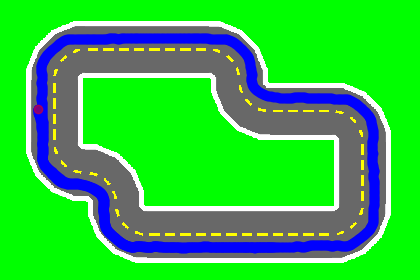}
		\caption{}
		\label{fig:sim-map-a}

	\end{subfigure}
	\begin{subfigure}{0.2\textwidth}
		\centering
		\includegraphics[height=2.3cm]{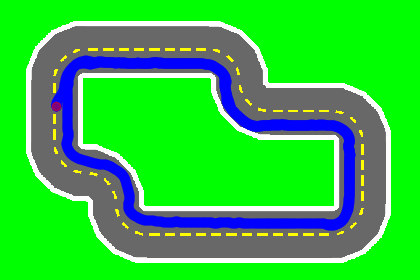}
		\caption{}
		\label{fig:sim-map-b}
	\end{subfigure}
	\\
	\par\medskip
	\begin{subfigure}{0.2\textwidth}
		\centering
		\includegraphics[height=2.3cm]{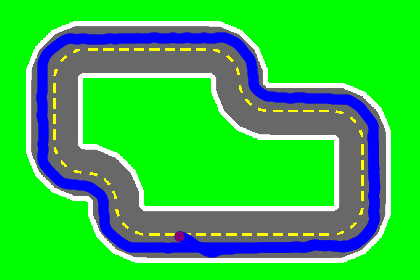}
		\caption{}
		\label{fig:sim-map-c}
	\end{subfigure}
	\begin{subfigure}{0.2\textwidth}
		\centering
		\includegraphics[height=2.3cm]{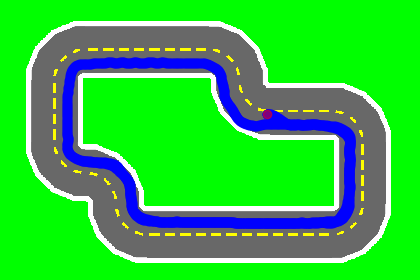}
		\caption{}
		\label{fig:sim-map-d}
	\end{subfigure}
	\\
	\par\medskip
	\begin{subfigure}{0.2\textwidth}
		\centering
		\includegraphics[height=2.3cm]{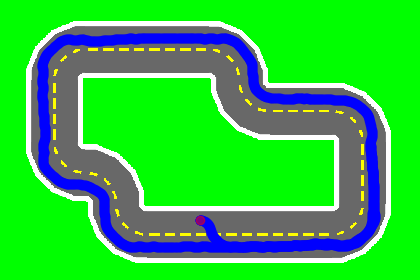}
		\caption{}
		\label{fig:sim-map-e}

	\end{subfigure}
	\begin{subfigure}{0.2\textwidth}
		\centering
		\includegraphics[height=2.3cm]{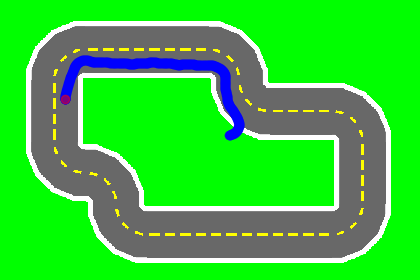}
		\caption{}
		\label{fig:sim-map-f}

	\end{subfigure}
	\\
	\par\medskip
	\begin{subfigure}{0.2\textwidth}
		\centering
		\includegraphics[height=2.5cm]{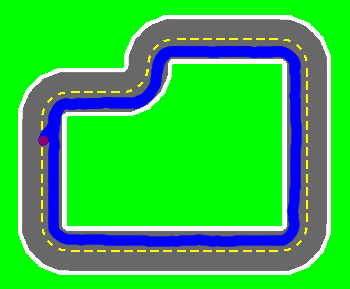}
		\caption{}
		\label{fig:sim-map-g}

	\end{subfigure}
	\begin{subfigure}{0.2\textwidth}
		\centering
		\includegraphics[height=2.5cm]{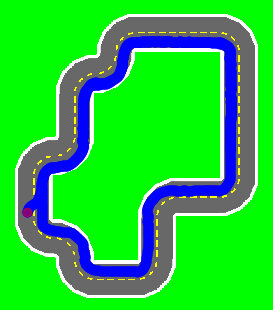}
		\caption{}
		\label{fig:sim-map-h}

	\end{subfigure}
	\\
	
	\caption{The paths of the robot on the maps after starting it from different locations in the simulator. The initial location of the robot is marked red. \ref{fig:sim-map-a}-\ref{fig:sim-map-e}, \ref{fig:sim-map-g}, \ref{fig:sim-map-h} show attempts where the robot was able to drive successfully a complete lap, with \ref{fig:sim-map-e} showing a situation where the robot was started from an invalid position but still was able to go to the right lane. \ref{fig:sim-map-f} shows a scenario where the robot failed to take a whole lap. \ref{fig:sim-map-a}-\ref{fig:sim-map-f}, \ref{fig:sim-map-g} and \ref{fig:sim-map-h} show Map \#1, Map \#2 and Map \#3, respectively.}
	\label{fig:fig-maps}
\end{figure}

\begin{figure}[tb]
	\centering
	\begin{subfigure}{0.2\textwidth}
		\centering
		\includegraphics[height=2cm]{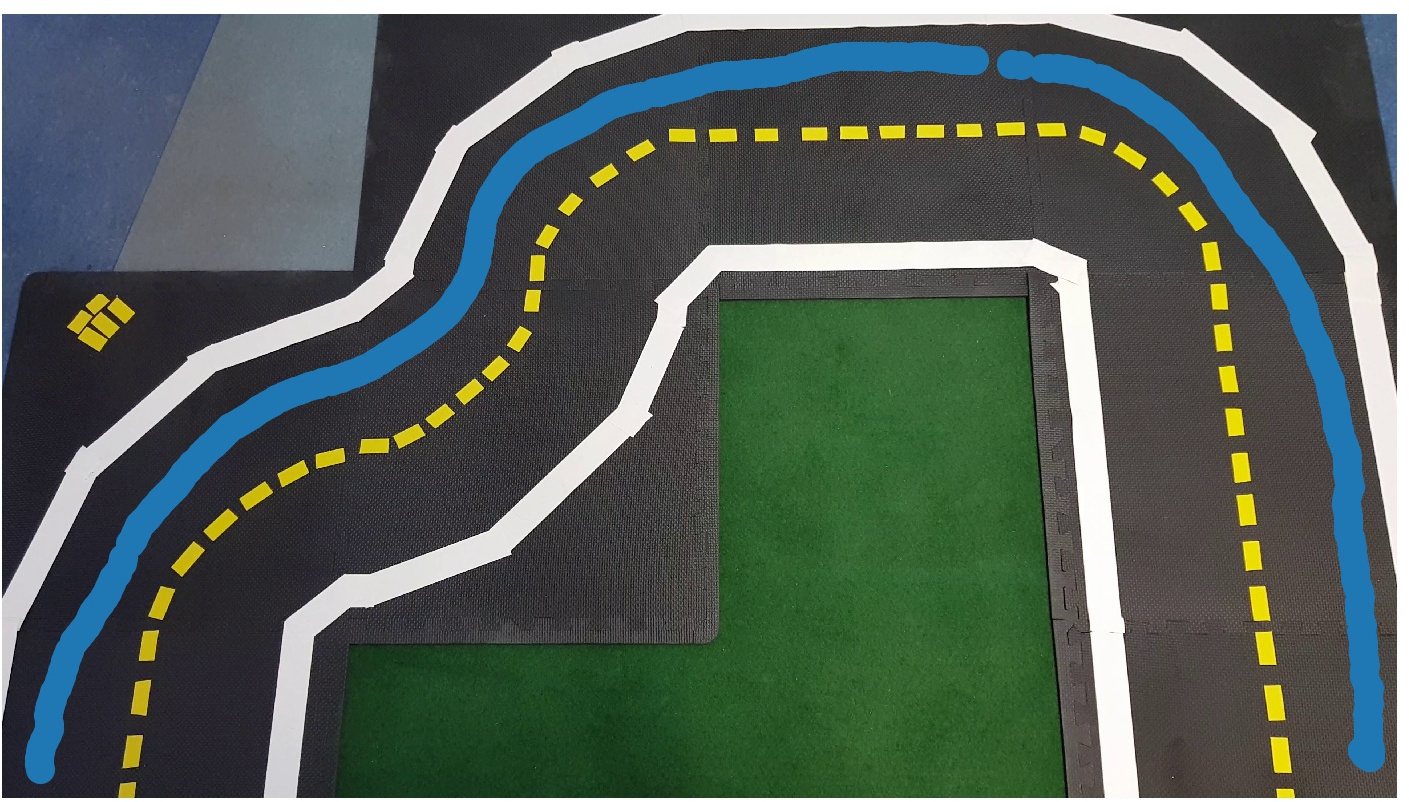}
		\caption{}
		\label{fig:sim-rmap-a}

	\end{subfigure}
	\begin{subfigure}{0.2\textwidth}
		\centering
		\includegraphics[height=2cm]{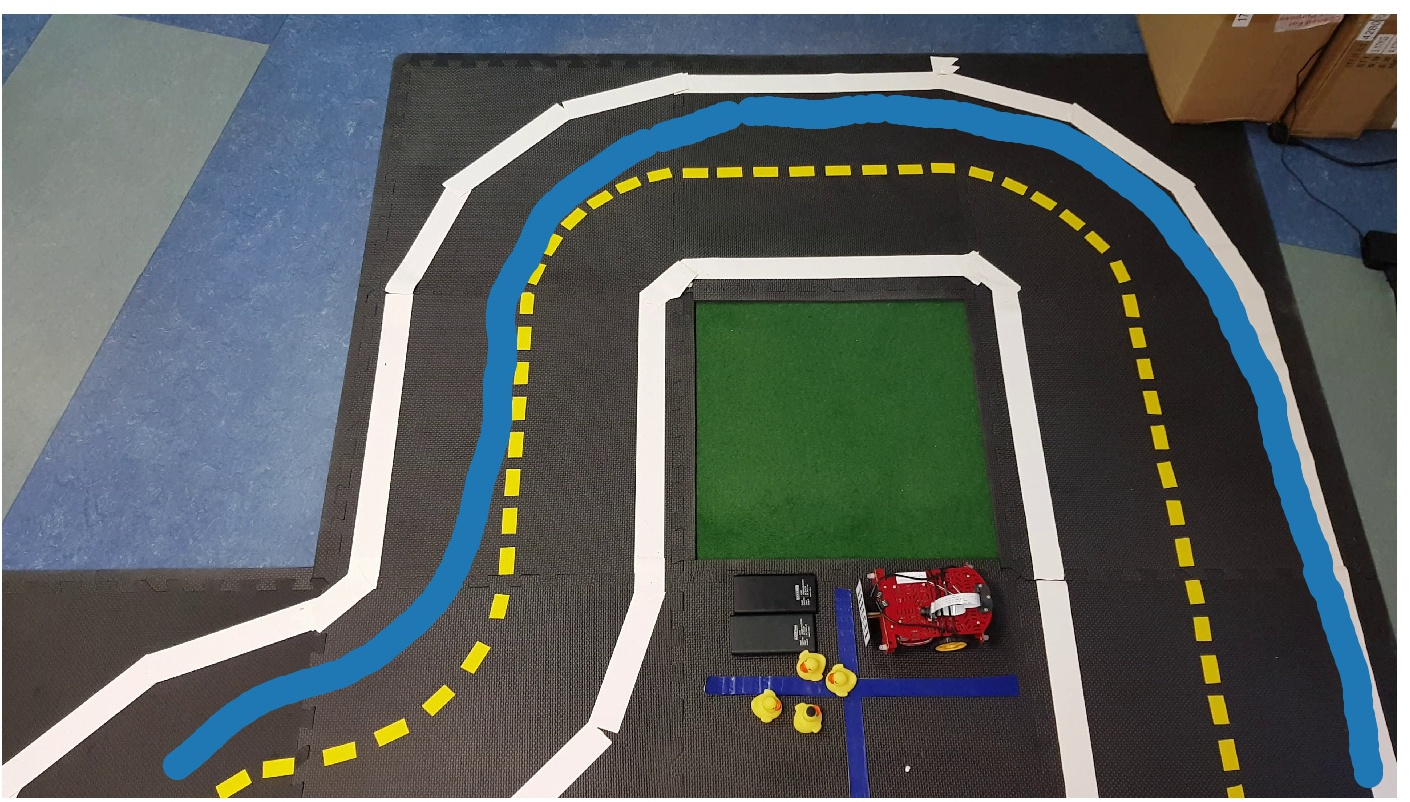}
		\caption{}
		\label{fig:sim-rmap-b}
	\end{subfigure}
	\\
	\par\medskip
	\begin{subfigure}{0.2\textwidth}
		\centering
		\includegraphics[height=2cm]{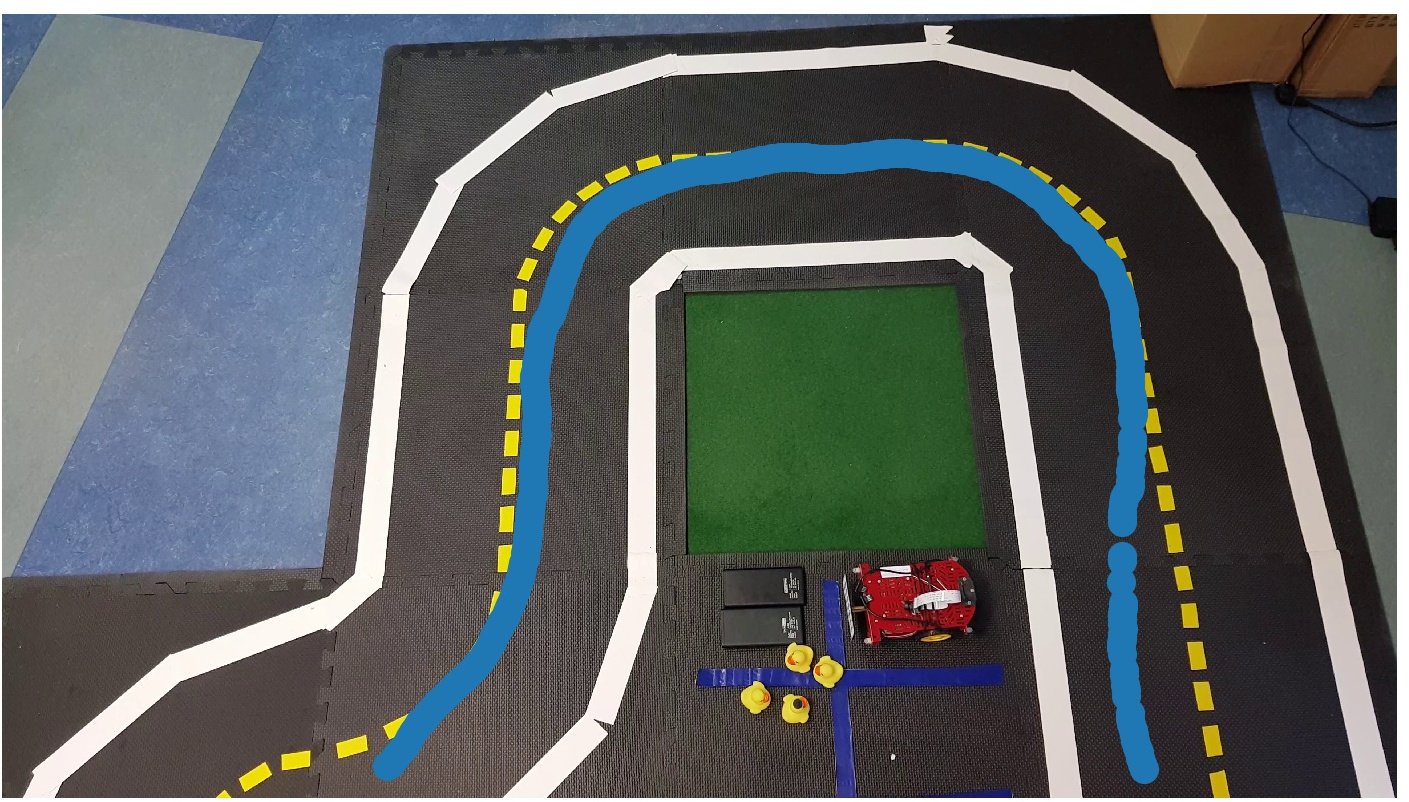}
		\caption{}
		\label{fig:sim-rmap-c}
	\end{subfigure}
	\begin{subfigure}{0.2\textwidth}
		\centering
		\includegraphics[height=2cm]{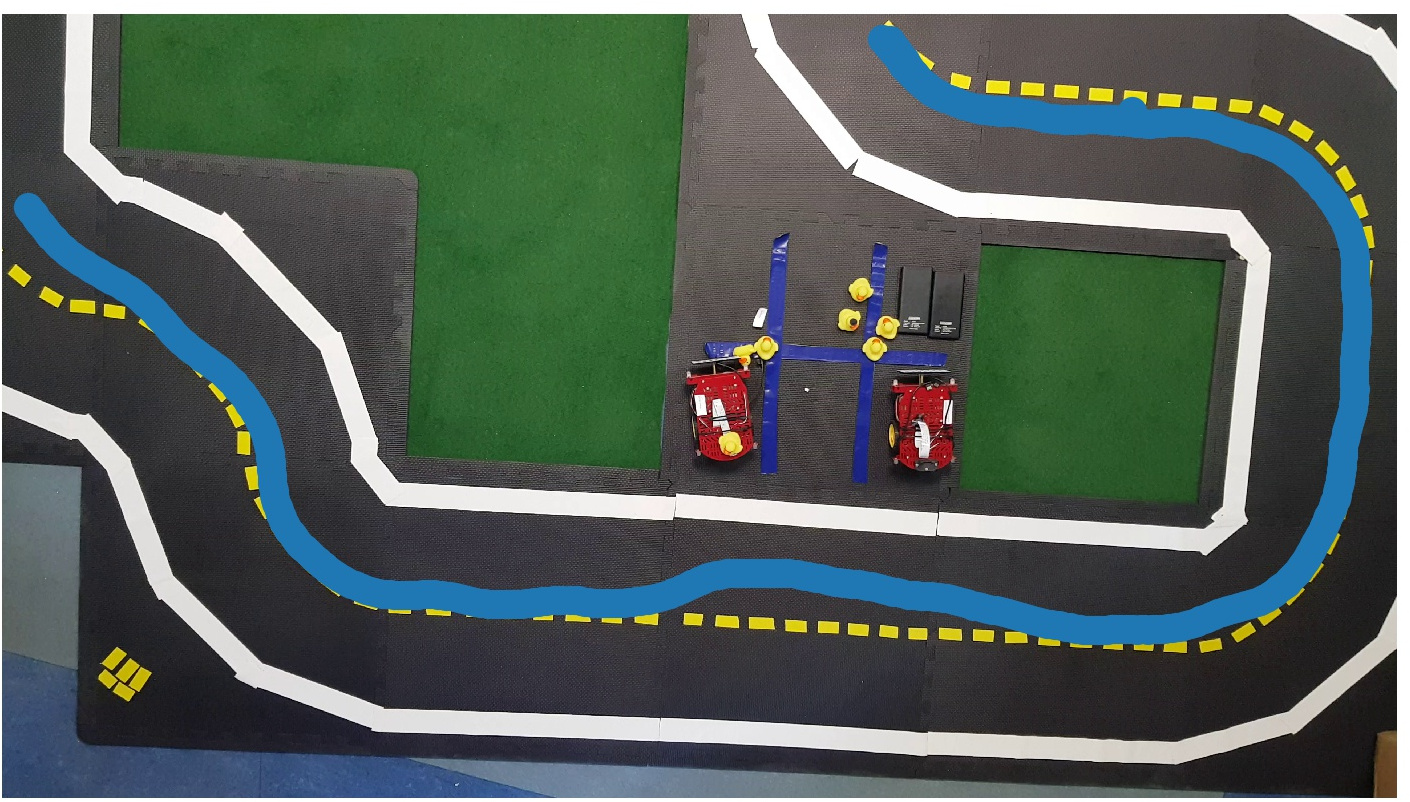}
		\caption{}
		\label{fig:sim-rmap-d}
	\end{subfigure}
	\\

	\caption{The paths of the real robot on the map after starting it from different locations in the real-world environment.}
	\label{fig:fig-real-maps}
\end{figure}

Fig. \ref{fig:fig-maps} shows the paths of the robot after starting it from various randomized locations in the simulator. The robot was able to navigate to the center of its lane and drive a whole lap there even when it was started from an invalid location, the oncoming lane (see \ref{fig:sim-map-e}). Fig. \ref{fig:fig-real-maps} shows the paths of the real-world robot. To created these images, we placed an ArUco marker \cite{garrido2016generation} \cite{romero2018speeded} on the top of the robot. We created a plan view video of the evaluation with a fixed camera and ran the ArUco detector algorithm for each frame of the video. The found locations of the marker are drawn on the map. Although occasionally the robot touched the middle yellow line during the evaluation, mostly it successfully ran in its lane. (As the camera was not completely above the robot all the time, the images have a small distortion: the marker on the top of the robot is not projected directly under the center of the robot due to the angle between the robot and the plan view camera.)

\subsubsection{Analysis of the predictions}

\begin{figure}[tb]
		\centering
		\begin{subfigure}{0.48\textwidth}
			\centering
			\includegraphics[height=4.5cm]{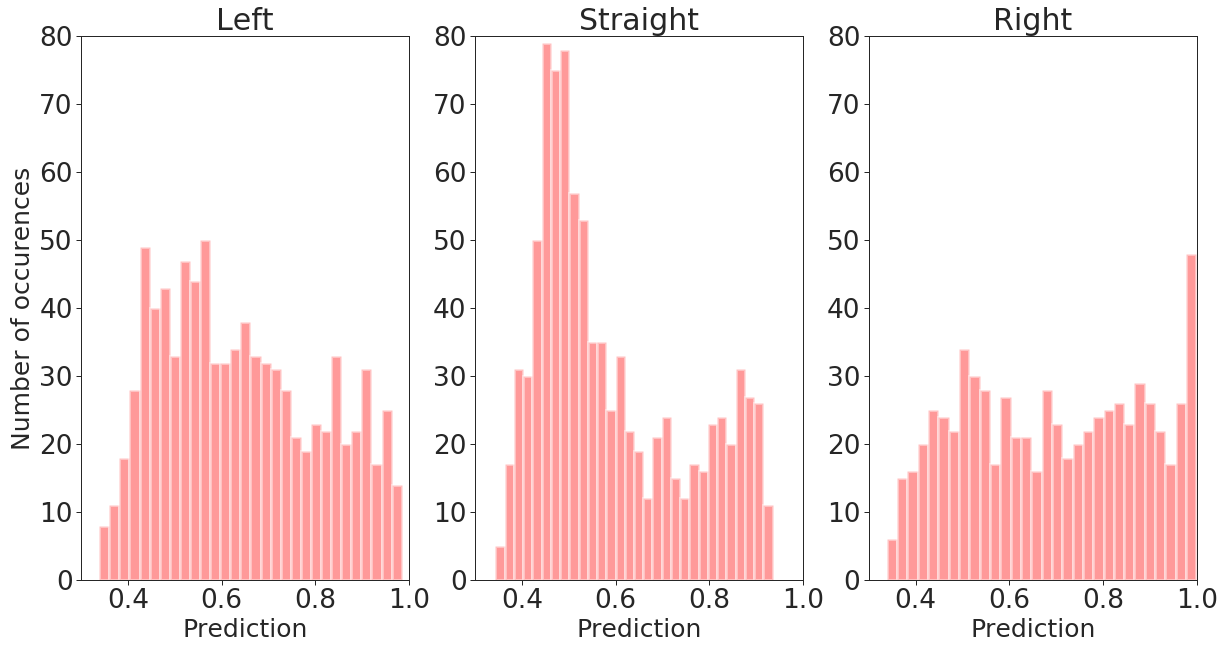}
			\caption{Simulator}
			\label{fig:fig-probs-sim}
		\end{subfigure}
		\begin{subfigure}{0.48\textwidth}
			\centering
			\includegraphics[height=4.5cm]{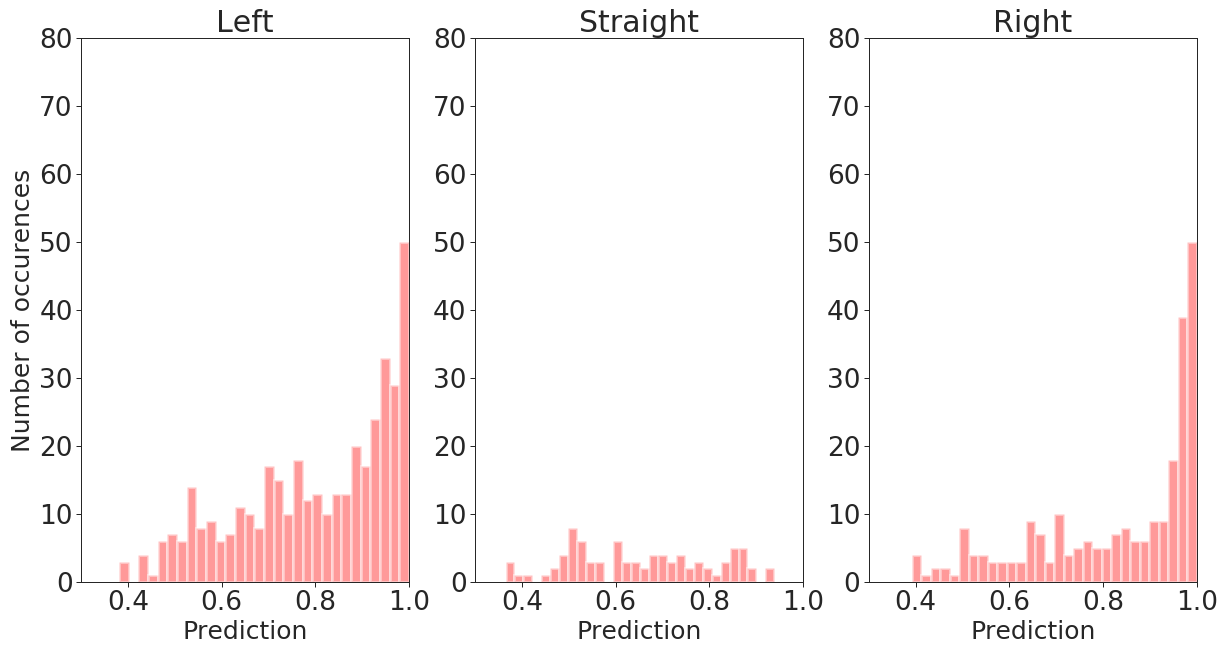}
			\caption{Real world}
			\label{fig:fig-probs-real}
		\end{subfigure}
		\caption{Histogram of the predictions of the model in the simulator (\ref{fig:fig-probs-sim}) and in the real world (\ref{fig:fig-probs-real}), separated based on the predicted action.}
		\label{fig:fig-probs}

	\end{figure}

Fig. \ref{fig:fig-probs-sim} shows the histogram of the probabilities predicted by the model for the selected action in the simulator, separated by the predicted actions. Fig. \ref{fig:fig-probs-real} shows the same metrics for the real world. Both histograms are made from the predictions generated while driving one lap on the track. We assume that the differences are due to the physical aspects of running the agent in the real world. To predict the \textit{Straight} action, the robot has to be approximately in the middle of the lane, facing forward (otherwise, it can get closer to the middle of the lane by turning, and thus can get a higher reward). Positioning to the middle of the lane is easier in the simulator, as the agent generates predictions and gives commands more frequently (around 2500 times in the simulator and 700 times in the real world during one whole lap), which means that more corrections are required in the real world to stay in the lane. Also, the control commands have an unsteady delay in the real world consisting of the network delay and inference time, which results in the need for more corrections accomplished by generating turning commands. 

\subsubsection{Comparing with state-of-the-art results}

The AI Driving Olympics (AI-DO) is a series of competitions focusing on AI for self-driving vehicles in the Duckietown environment \cite{zilly2019ai}. The competition had three rounds in 2018 and 2019 organized at the NeurIPS and ICRA conferences. The goal of the competition is to make it possible to test the recent theoretical advances in the area in practice. Different tasks can be solved at the competition: the simplest is lane following, but more complex ones, such as navigating in the presence of other vehicles and handling intersections, are also available. In the case of lane following, the task is to process the image of the robot's camera and give wheel speed commands based on it to navigate the robot on the map. The competition makes it possible to compare different methods and algorithms and evaluate their performance.

We compared our method to the ones that were used by the winners of the 3rd AI Driving Olympics. In the competition, the submissions were tested on a real robot for 30 seconds. The goal was to drive as far as possible without leaving the track in the given time limit. The driven distance was discretized to the number of map tiles the robot successfully passed. The submissions were tested from two different initial locations. While we are not able to test our method on the same track, the competition track and ours are built from the same standardized elements (straight roads and turns). We performed the tests by starting our robot from two different initial locations for 30 seconds, and measured the distance it covered. The results can be seen in Table \ref{tab:aido}. The top submissions in the competition mostly used imitation learning-based methods in contrast to our reinforcement learning-based approach. It is worth noting that while our method is currently not optimized for driving the robot as fast as possible, it has comparable performance to the state-of-the-art solutions.

\begin{table}[tbp]
\caption{Results of the best performing, state-of-the-art agents in AI-DO 3 competition and our approach.}
\begin{center}
\begin{tabular}{lSS}
\toprule
\multirow{2}{*}{\textbf{Team}} & \multicolumn{2}{c}{\textbf{Distance driven (tiles)}} \\
& \textbf{Run \#1} & \textbf{Run \#2} \\
\midrule
\textbf{JBRRussia1} & 11 & 19 \\
%\hline
\textbf{phmarm} & 10 & 18 \\ 
%\hline
\textbf{JBRRussia} & 8 & 2 \\ 
%\hline
\textbf{miksaz} & 8 & 1 \\ 
\midrule
\textbf{Our approach} & 12 & 13 \\
\bottomrule

\end{tabular}
\label{tab:aido}
\end{center}
\end{table}

\subsubsection{Other experiments}

We ran several experiments to find the best training parameters and image preprocessing method. In the following, we present our experiences regarding the use of each image preprocessing step. \textit{Resizing} the image is required to make it possible to control the robot in real-time with as little latency as possible. We used a single laptop with no dedicated graphics card and a 4-core Intel\textregistered Core\texttrademark i7-4500U CPU @ 1.80GHz. We measured that creating one prediction takes approx. 3-4 milliseconds, which is adequate with the camera frames arriving at a rate of 30fps. \textit{Cropping} the upper part of the images helps to transfer to the real robot, as this step eliminates most of the objects around the track that would otherwise fall into the field of view of the robot and thus could make navigating more difficult. The \textit{color segmentation} also improves the transfer to the real robot, as it highlights the important parts of the image and hides the objects surrounding the track. \textit{Normalization} is required for the more effective training of the neural network. The \textit{image buffer} has an important role in stabilizing the movement of the robot. Without this, the agent usually navigates in a straight line by alternating between actions 0 and 1 (according to Table \ref{tab:discrete-actions}), which results in an oscillating movement. Using the image buffer helps the robot to find the center of its lane and go straight there; while the oscillating movement sometimes still occurs, it is much less frequent than in the absence of the image buffer. The oscillating movement of the robot can be smoothed by using a larger number of discrete actions (e.g. 5 or 7), however, it makes training the agent more difficult.

We tested the robustness of our method by running experiments in real-world environment in different lighting conditions and robot speeds. We changed the lighting conditions by varying the number and position of the lights turned on in the evaluation room. We changed the speed of the robot by multiplying the original speed values with constants, thus, making the general movement of the robot slower or faster. We found that these changes had no effect on the performance of the robot, i.e. it produced similar performance with and without these changes.

%TODO itt legyen meg valami (pl. egyeb metrikak?)

\section{Conclusions}
\label{sec:conclusion}

The difference between the simulator and the real world is a major challenge for applications of reinforcement learning in robotics and autonomous driving. In this paper, we presented a pipeline for a Deep Reinforcement Learning-based algorithm to perform autonomous robot control using Deep Q-Networks. We proposed a method to train the agent in a simulator, which can later control the robot both in the simulated and the real-world environment. We used the Duckietown environment to evaluate our method. We showed that using the proposed approach, the trained model is capable of navigating the robot along the track both in the simulator and the real world. Our method has comparable performance to the state-of-the-art solutions and can be run real-time on limited hardware resources.

\section*{Acknowledgements}
The research presented in this paper has been supported by Continental Automotive Hungary Ltd., by the European Union, co-financed by the European Social Fund (EFOP-3.6.2-16-2017-00013), by the BME-Artificial Intelligence FIKP grant of Ministry of Human Resources (BME FIKP-MI/SC). Bálint Gyires-Tóth is supported by the Doctoral Research Scholarship of Ministry of Human Resources (ÚNKP-19-4-BME-189) in the scope of New National Excellence Program and by János Bolyai Research Scholarship of the Hungarian Academy of Sciences. Péter Almási expresses his gratitude for the financial support of the Nokia Bell Labs Hungary.

%TODO hivatkozasokat atnezni

\printbibliography

\end{document}